\documentclass{article}
\PassOptionsToPackage{numbers, sort&compress}{natbib}
\usepackage{bbm}
\usepackage{amsthm}
\usepackage{amsmath}
\usepackage{amssymb}
\usepackage{color}
\usepackage{xspace}
\usepackage{pgfplots, pgfplotstable}
\usepackage{rotating}
\usepackage{multirow}
\usepackage{enumitem}

\def\hyper{{hyperparameter}\xspace}
\def\hypers{{hyperparameters}\xspace}

\newcommand{\comment}[1]{}

\newcommand{\argmax}[1]{\underset{#1}{\operatorname{argmax}}}

\newcommand{\R}[1]{R}
\newcommand{\G}[1]{G}

\newcommand{\Ad}[1]{A}

\newcommand{\CC}[1]{C}

\def\eg{{\em e.g.,}}
\def\ie{{\em i.e.,}}

\def\vs{{\em vs.}\xspace}

\newcommand{\figref}[1]{Figure~\ref{#1}}
\newcommand{\tabref}[1]{Table~\ref{#1}}

\usepackage[square,numbers]{natbib}
\usepackage[final]{nips_2018}
\usepackage[utf8]{inputenc} 
\usepackage[T1]{fontenc}    
\usepackage{hyperref}       
\usepackage{url}            
\usepackage{booktabs}       %
\usepackage{algorithm}
\usepackage{algorithmic}
\usepackage[titletoc,title]{appendix}
\usepackage{bbm}

\usepackage{amsfonts}       
\usepackage{nicefrac}       
\usepackage{microtype}      
\title{Parallel Architecture and Hyperparameter Search\\via Successive Halving and Classification}

\author{
\begin{tabular}{c@{\hspace*{.9cm}}c@{\hspace*{.9cm}}c@{\hspace*{.9cm}}c}
Manoj Kumar\thanks{Work done as part of the Google AI Residency Program.} &
George E.~Dahl &
Vijay Vasudevan &
Mohammad Norouzi \\
\end{tabular}\\[.1cm]
\texttt{\{mechcoder, gdahl, vrv, mnorouzi\}@google.com}\\
Google Brain
}

\begin{document}
\maketitle
\begin{abstract}

We present a simple and powerful algorithm for parallel black box
optimization called Successive Halving and Classification (SHAC).  The
algorithm operates in $K$ stages of parallel function evaluations and
trains a cascade of binary classifiers to iteratively cull the
undesirable regions of the search space.  SHAC is easy to implement,
requires no tuning of its own configuration parameters, is invariant
to the scale of the objective function, and can be built using any
choice of binary classifier. We adopt tree-based classifiers within
SHAC and achieve competitive performance against several strong
baselines for optimizing synthetic functions, hyperparameters, and
architectures.

\comment{
\textcolor{red}{Finally, we study the problem of two-stage black box
optimization, where a shortlist of candidate points is selected based
on a proxy objective to facilitate a single round of parallel
evaluation on the final objective. We find that SHAC exhibits a high
degree of diversity in the selected shortlist, which is favorable when
there is a weak correlation between the proxy and final objective
functions.}
}


\end{abstract} 

\vspace*{-.2cm}
\section{Introduction}
\vspace*{-.1cm}

Artificial neural networks have seen success in a variety of application domains such as speech recognition~\cite{speech},
natural language processing~\cite{attentionmt}, and computer vision~\cite{alexnet}. 
Recent advances have come at the cost of a significant increase in the complexity of the neural systems.
State-of-the-art neural networks make use of many layers~\cite{resnet}, multiple branches~\cite{inception},
complicated connectivity patterns~\cite{densenet}, and different attention mechanisms~\cite{attentionmt,transformer},
in addition to tricks such as dropout~\cite{dropout} and batch normalization~\cite{batchnorm}.
Domain experts continue to develop new neural network practices,
sometimes resulting in improved models across domains,
but designing new architectures is time consuming and expensive.
To facilitate easier and faster development of the next generation of neural networks,
we need automated machine learning algorithms for tuning \hypers (\eg~\cite{bergstra2011algorithms,snoek2012,snoek2015scalable,pbt})
and selecting architectures (\eg~\cite{nas,baker2016designing,real2017,liu2017hierarchical}).

Hyperparameter and architecture search are instances of {\em black box optimization}, where one seeks the
maximum (or minimum) of a function not available in closed form via iterative evaluation on a small number of proposed candidate points.
A few crucial characteristics of a successful and practical black box optimization algorithm are:
\vspace*{-.1cm}
\begin{enumerate}
    \item {\em Robustness}: The algorithm should require no tuning to achieve stable maxima (or minima) across
          different domains with different attributes and evaluation budget requirements.
    \item {\em Parallelism}: The algorithm should support
  parallel generation of candidate points to speed up optimization.  
    \item {\em Scalability}: The algorithm should scale to high dimensional search spaces.
\end{enumerate}
\vspace*{-.1cm}
Despite the introduction of many black box optimization algorithms in the past decade,
most practitioners continue to resort to random search because of its simplicity, robustness, parallelism, and scalability.
Empirical studies on multiple domains suggest that random search using a budget of twice as many point evaluations
outperforms many complex black box optimization algorithms~\cite{hyperband}.

Inspired by the success and simplicity of random search,
we aim to achieve a constant factor improvement over 
this baseline by iteratively culling the undesirable regions of the search space.
We propose an algorithm called {\em successive halving and classification (SHAC)},
which adopts a cascade of binary classifiers to evaluate the quality of different regions in the search space in a progressive manner.
To propose candidate points, points are randomly generated from the search space. Each classifier then filters the incoming points approved by the previous classifiers
and passes along half of the input points to the next classifier.
After a cascade of $k$ classifiers, we are left with a volume equal to $1/2^k$th of the original search space.
SHAC exhibits no preference between the candidate points that make it past all of the classifiers.
To select a new candidate point from the surviving region of the search space,
SHAC thus uses random search in the remaining volume by resorting to rejection sampling.

The SHAC algorithm is easy to implement, accommodates parallel point generation in each stage, and requires
almost no \hyper tuning, making it an excellent baseline for black box optimization research and a useful tool for practitioners.
SHAC is invariant to the scale of the evaluation metric and can support any binary classifier.
Unlike previous work that uses neural networks to design new neural networks, with the inner loop of SHAC
we recommend using classifiers that are simpler to train, easier to completely automate and that produce relatively consistent results on new problems; specifically, we suggest using gradient boosted trees~\cite{gboosting,xgboost}.
In practice, SHAC maintains a high degree of diversity among the candidate points it proposes,
which as we discuss later, is an essential feature when dealing with noisy measurements and unfaithful proxy tasks.

We conduct extensive empirical evaluations comparing SHAC with random
search and NAS~\citep{nas, nas2} on CIFAR-10 and CIFAR-100 on both
architecture and \hyper search and on the optimization of synthetic
functions.  Our experiments confirm that SHAC significantly
outperforms RS-2X: a Random Search baseline with twice as many point
evaluations across domains.  Importantly, SHAC outperforms
NAS~\cite{nas2} in the low data regime on hyper-parameter tuning.

\vspace*{-.2cm}
\section{Related work}
\vspace*{-.1cm}

{\bf Hyperparameters.}~~There has been a large body of previous work on automated \hyper tuning for neural networks.
\citet{rs} demonstrate that random search is a competitive baseline, often outperforming grid search.
{\em Bayesian optimization} techniques learn a mapping from
the \hypers to the validation scores using a surrogate model such as
Parzen window estimators~\cite{bergstra2011algorithms}, Gaussian
Processes~\cite{snoek2012}, Random Forests~\cite{smac} or even other
neural networks~\cite{snoek2015scalable}.  Such methods alternate
between maximizing an acquisition function to propose new \hypers and
refining the regression model using the new datapoints, \ie~updating
the posterior.  Another class of \hyper tuning algorithms performs
implicit gradient descent on {\em
continuous} \hypers~(\eg~\cite{bengio2000gradient,maclaurin2015gradient}).
By contrast, our approach casts black box optimization as iterative
classification rather than regression.  Our final goal is to find a
population of diverse \hypers that consistently perform well, rather
than finding a single setting of \hypers.

Our work is inspired by {\em Successive
Halving} \cite{successivehalving}, a population based \hyper tuning
algorithm.  Successive halving starts with a large population
of \hypers, \eg~$2^k$ instances, and iteratively prunes them by
halving the population at every stage, \eg~retaining $1$ \hyper out of
$2^k$ after $k$ stages.  At each stage, Successive Halving trains the
models in the population for some additional number of steps and only
retains the models that outperform the population median at that
stage.
\citet{hyperband} suggest a scheme for balancing the number of initial \hypers with
the amount of resources allocated for each \hyper setting and shows
some desirable theoretical guarantees for the algorithm.  Recently,
similar population based techniques~\cite{pbt} have become popular for
tuning the \hypers of reinforcement learning algorithms.  Our method
is similar to Successive Halving, but we rely on a classifier for
pruning the points at each stage, which significantly reduces the
computational cost associated with the optimization algorithm,
especially in the initial stages of the search.  Furthermore, while
Successive Halving only applies to the optimizatifon of iterative
machine learning models, our proposed technique is a generic black box
optimization algorithm.

Recently, \citet{hashimoto2018} independently developed an iterative
classification algorithm for derivative free optimization inspired by
cutting plane algorithms~\cite{nesterov2013intro}. One can think of
their proposed algorithm as a {\em soft} variant of SHAC, where
instead of making hard decisions using a classification cascade, one
relies on the probability estimates of the classifiers to perform soft
pruning. They theoretically analyze the algorithm and show that given
sufficiently accurate classifiers, one can achieve linear convergence
rates. We leave comparison to soft classification to future work, and
focus on large-scale experiments on hyperparameter and architecture
search.


{\bf Architectures.}~~There has been a surge of recent interest in
automating the design of neural networks.  The distinction between
architectures and hyperparameters is mainly qualitative, based on the
size and the expressiveness of the search spaces. Hyperparameter
spaces tend to be smaller and well-specified, whereas architecture
spaces tend to be vast and ill-defined.  Specifying an expressive
encoding of architectures that can easily represent successful
architectures,
\eg~different convolutional networks~\cite{alexnet,vgg,inception},
is an important research problem in its own right.  One family of
approaches develop fixed length codes to represent reusable
convolutional blocks for image
recognition~\cite{zhong2017practical,nas2}.  Another family focuses on
evolutionary algorithms and mutation operations that iteratively grow
a graph representation of the
architectures~\cite{real2017,liu2017hierarchical,miikkulainen2017evolving,liu2017progressive}.
Unfortunately, direct comparison of different architecture search
techniques is challenging because they often use different search
spaces.  Even on the same search space, methods with different
computational budgets are difficult to compare.  The computational
issue is more subtle than simply counting the total number of
architectures tested because testing $n$ architectures in parallel is
not the same as testing $n$ architectures sequentially.  Moreover, one
may abandon architectures that seem ineffective in the early phases of
training.  One natural way of expressing a computational budget is
with a maximum number of parallel workers and a maximum number of
total time steps.  In our experiments, we compare different
architecture search algorithms that use the same search spaces.  We
give each algorithm access to an equal number of parallel workers
within each round and an equal number of rounds.  We replicate the
experimental setup and the search space of \cite{nas2} with minor
changes, and we train all of the architectures for a fixed number of
epochs.

Previous work applies different black box optimization algorithms to architecture search.
\citet{nas} and \citet{baker2016designing} cast the problem as reinforcement learning~\cite{suttonbook}
and apply policy gradient and Q-learning respectively.
\citet{pnas} use a surrogate RNN as a predictive model to search for architectures with increasing complexity.
\citet{deeparchitect} use Monte Carlo Tree search, while other papers adopt ideas from {\em neuroevolution}~\cite{miller1989designing, holland1992adaptation, stanley2002evolving}
and apply evolutionary
algorithms~\cite{real2017,liu2017hierarchical,miikkulainen2017evolving}
to architecture search.
\citet{smash} learn a hypernetwork~\cite{hypernetworks} to predict the weights of a neural network given the architecture for fast evaluation.
\citet{baker2017} suggests learning a mapping from architectures and initial performance measurements to the corresponding final performance.

In the face of the inherent uncertainty and complexity of empirically
evaluating architecture search techniques, we advocate using simple
algorithms with few of their own \hypers that achieve competitive
accuracy across different domains and computation budgets.  Extensive
tuning of a given search algorithm for a particular search space might
achieve impressive performance, but all the computation to tune the
search algorithm must be counted against the computation budget for
the final run of the algorithm. Benchmarks and empirical evaluation of
architecture search methods have not yet progressed enough to risk
intricate algorithms that themselves require substantial tuning.


\section{SHAC: Successive Halving and Classification}

The task of black box optimization entails finding an {\em
  approximate} maximizer of an objective function $b(x)$ using a total
budget of $N$ point evaluations\footnote{For architecture and
  hyperparameter search, we are interested in maximizing the black-box
  objective for e.g.\ the mean cross-validation accuracy which is why
  we denote this as a maximizer instead of a minimizer}.
\begin{equation}\label{eq:maxa}
\widehat{x}^* ~\approx~ x^* ~=~ \argmax{x}~b(x)~.
\end{equation}
Typical black box optimization algorithms alternate between evaluating
candidate points and making use of the history of previous function
evaluations to propose new points in the promising and unknown regions
of the search space (\ie~explore exploit dilemma). A good black box
optimization algorithm is expected to find an approximate maximizer
$\widehat{x}^*$ that approaches $x^*$ in the limit of infinite
samples, and there is a notion of {\em asymptotic regret} that
captures this intuition. However, in most practical applications one
has to resort to empirical evaluation to compare different algorithms
at different budgets.

In this paper, we study \emph{parallel} black box optimization, where
the budget of $N$ points is divided into $m$ batches where each batch
can be evaluated using $W = {N}/{m}$ workers in parallel. In this
setup, the optimization algorithm should facilitate parallel
generation and evaluation of a batch of $W$ candidate points
$\{x_i^{(j)}\}_{i=1}^W$ in the $j$th round to make use of all of the
available resources. When $m$ is small, random search is one of the
most competitive baselines because more sophisticated algorithms have
very few opportunities to react to earlier evaluations. Using SHAC, we
aim to get a constant factor improvement over RS, even when $m$ is
not large.

SHAC uses a cascade of $k$ binary classifiers denoted $(c_1(x), \dots,
c_{k}(x))$ to successively halve the search space. Let the output of
each classifier denote the predicted binary label, \ie~$c_i(x) \in
\{-1, 1\}$. To propose a new candidate point, SHAC generates a random
point from the prior distribution and rejects it if any of the
classifiers predict a negative label, \ie~a point is rejected if
$\exists i\in \{1,\ldots,k\}$ such that $c_i(x) = -1$. Given binary
classifiers that on average reject $50\%$ of the incoming points, this
procedure amounts to accepting a volume of about ${1}/{2^k}$th of the
search space. We train the $(k\!+\!1)$th binary classifier,
$c_{k+1}(x)$, on the population of points that make it past all of the
$k$ previous classifiers. Once all of the points in this set are
evaluated, we find the median function value of the population, and
assign a positive label to points with a value of $b(x)$ above the
median and a negative label otherwise. Once the $(k\!+\!1)$th
classifier is trained, it becomes the final classifier in the cascade
and is used along with the previous classifiers to generate the next
batch of points and repeat the process. The loop of proposing new
candidates and training new classifiers continues until we reach a
maximum number of allowed classifiers $K$. After $K$ classifiers have
been trained, the classifier cascade is frozen and used only to
generate any remaining points before exhausting the total budget of
$N$ evaluations. See Algorithm~\ref{alg:shac} for the pseduocode of
the SHAC algorithm.

SHAC is easy to implement and use. By casting the problem of black box
optimization as iterative pruning using a cascade of binary
classifiers, we conservatively explore the high performing regions of
the space. SHAC requires no tuning of its own configuration parameters
and is invariant to the scale of the objective function. We discuss
the configuration parameteres of SHAC below.

\begin{figure*}[t]
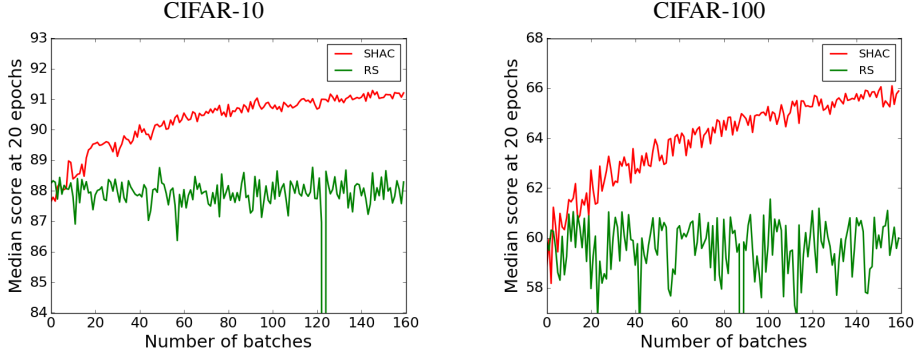

{
\small
\begin{center}
\begin{tabular}{@{}c@{\hspace*{1cm}}c@{\hspace*{.3cm}}c@{}}
CIFAR-10 & CIFAR-100 \\
\includegraphics[width=.4\textwidth]{cifar10_median.pdf}
&
\includegraphics[width=.4\textwidth]{cifar100_median.pdf}
&
\end{tabular}
\end{center}
}
\caption{
    We compare the median function values of every batch of points proposed by RS and SHAC on CIFAR10 and CIFAR100 architecture search problems (see Section~\ref{arch_proxy}).
    Not surprisingly, the median score of the points proposed by RS fluctuates around a constant. As expected, the median scores of the points proposed by SHAC fluctuate around a monotonically increasing curve.}
\label{fig:cifar_median}
\end{figure*}

\textbf{Binary classifiers:}~Within the SHAC algorithm, any family of binary classifiers can be used. In our experiments, we use gradient boosted trees~\cite{gboosting} to classify points at every stages. Gradient boosted trees have shown to be flexible, fast (for both training and inference), robust to the choice to their own hyperparameters, and easy to get working in new problem domains~\cite{xgboost}. These characteristics are all helpful properties of any black box optimization algorithm and accordingly we recommend using gradient boosted trees or random forests~\cite{rf} within the SHAC algorithm. In all of our experiments, we use gradient boosted trees, fix the number of trees to be $200$ and do not change any of the other default hyperparameters in the XGBoost implementation \cite{xgboost}. It is expected that increasing the number of trees is likely to improve the performance at the cost of some computation overhead.

\textbf{Maximum number of classifiers:}~If we train and adopt a classifier after every batch of points is evaluated, then we will end up with a maximum of $m - 1$ classifiers.
Given $K$ classifiers, to draw a new point that makes it past all of the classifiers, one needs to draw on average $2^K$ random points, one of which will be approved.
In order to reduce the computational overhead, we limit the number of classifiers to a maximum of $18$ and define the number of classifiers as $K=\min(m-1, 18)$.

\textbf{Classifier budget:} \comment{ When the maximum number of
  classifiers $K$ is small compared to the budget (in other words, $KW
  < N$), classifiers might be trained on a very small number of points
  and not make useful accept/reject decisions.}  Given a budget of $N$
point evaluations and a total of $K$ classifiers, it is natural to
distribute the budget evenly among the classifiers, so each classifier
is trained on a reasonably sized dataset.  To accomodate the parallel
budget of $W$ workers, we set the minimum number of points per
classifier to $T_c = {W \lfloor \frac{N}{W(K + 1)} \rfloor}$ in all
our experiments. Further, to make sure that completely useless
classifiers are not used, we only adopt a new classifier if its
$5$-fold cross validation accuracy is at least $50\%$.

\begin{algorithm}[t]
  \caption{SHAC: Successive Halving and Classification}
  \label{alg:shac}
\begin{algorithmic} 
  \STATE {\bfseries Input:} prior distribution $U(x)$, black-box objective $b(x)$, total budget $N$, classifier budget $T_{c}$
   \STATE {\bfseries Initialize:} $C \leftarrow \{\}$ \comment{\hspace*{3cm}$\rhd$~$C$ is the set of binary classifers.}
   \STATE {\bfseries Initialize:} $S \leftarrow \{\}$ \comment{\hspace*{3.03cm}$\rhd$~$S$ is the current dataset of input output pairs.}
   \FOR {$t=1$ to $N$}
   \REPEAT
   \STATE Sample $x \sim U(x)$
   \UNTIL{$\forall c \in C:~c(x) > 0$} \comment{\hspace*{1.87cm}$\rhd$~Rejection sampling}
   \STATE Evaluate $y = b(x)$
   \STATE $S \leftarrow S \cup \{(x, y)\}$
   \IF{$|S| = T_{c}$}
       \STATE $\{(x_i, y_i)\}_{i=1}^{\lvert S \rvert} \leftarrow S$
       \STATE $\tilde{y} = \operatorname{median}(
    \{ y_i \}_{i=1}^{\lvert S \rvert})$
       \STATE $S' = \{(x_i, \mathrm{sign}(y_i - \tilde{y})) \}_{i=1}^{\lvert S \rvert}$\comment{\hspace*{.9cm}$\rhd$~Binarize the labels}
        \STATE Train a binary classifier $c$ on $S'$
        \STATE $C \leftarrow C \cup \{ c \}$
        \STATE $S \leftarrow \{\} $
   \ENDIF
   \ENDFOR
\end{algorithmic}
\end{algorithm}\vspace*{-.3cm}

\section{Experiments}
\vspace*{-.1cm}

To assess the effectiveness of the SHAC algorithm, we compare SHAC
with {\bf NAS{-PPO}}: Neural Architecture Search~\cite{nas2} based on
Proximal Policy Optimization (PPO)~\cite{ppo}, {\bf RS}: Random
Search, and {\bf RS-2X}: Random Search with twice the number of
evaluations. We conduct experiments on architecture search and
hyperparameter search on CIFAR-10 and CIFAR-100. We also run
experiments on two synthetic test functions used by previous work:
Branin and Hartmann6. The results for NAS are obtained with an
implementation generously provided by the respective authors and we
use the default configuration parameters of NAS. The NAS
implementation is an improved version of~\cite{nas} based
on~\cite{ppo}.

Since the entire search process, including the objective function, is
highly stochastic, the top point found by each algorithm varies quite
a bit across multiple runs. For architecture and hyperparameter
search, it is computationally prohibitive to run the search multiple
times, so we report the mean of the top $5$ values instead of the
single best result.

\subsection{Synthetic functions}

We adopt the Branin and Hartmann6 synthetic functions used by prior
work including~\cite{bergstra2011algorithms}. These functions present
accessible benchmarks that enable fast experimentation. For Hartmann6,
$x$ is a continuous $6$D vector, and each dimension has a uniform
prior in the interval $[0.0, 1.0]$. For Branin, $x$ is a continuous
$2$D vector, where the first and second dimensions have a uniform
prior in the range of $[-5.0, 10.0]$ and $[0.0, 15.0]$
respectively. Since these functions are available in closed form, it
is efficient to compute $b(x)$ at the proposed candidate points.

We compare SHAC to RS and RS-2X on a budget of $200$ and $400$
evaluations, where the budget is divided into batches of $20$ parallel
evaluations. Each classifier in SHAC is trained on a dataset of $20$
points. Since $20$ points are not enough to obtain a reliable
cross-validation estimate, we do not perform cross-validation here. We
report the empirical results in \tabref{tab:table1}.  We note that the
functions are being minimized here, so smaller numbers are
preferred. Because these experiments are cheap, we run each method $5$
times using different random seeds and report the mean and standard
error in \tabref{tab:table1}.  On both budgets on both functions, we
observe that SHAC significantly outperforms RS-2X. For comparison,
Spearmint~\cite{snoek2014input}, which uses a Gaussian Process for
black box optimization achieves $0.398$ and $-3.133$ on Branin and
Hartmann6 respectively using a sequence of $200$ function
evaluations. Spearmint outperforms SHAC at at the cost of fully
sequential evaluation, which is significantly slower in practice for
real world applications. SHAC on the other hand, leverages $20$
parallel evaluations in each step, and unlike Gaussian Processes
easily scales to very large datasets.

\vspace*{-.2cm}
\subsection{Hyperparameter Search}
\vspace*{-.1cm}

\begin{table*}[t]
{
\small
\begin{center}
    \begin{tabular}{| c | c | c | c | c | c | }
     \hline
     Dataset & (Batches, Workers) & RS & RS-2X & NAS{-PPO} & SHAC \\
     \hline
     \multicolumn{6}{|c|}{Synthetic functions} \\
     \hline
        Branin & 20, 20 & 0.543 $\pm$ 0.06 & 0.457 $\pm$ 0.01 & - & \textbf{0.410 $\pm$ 0.01}  \\
        \hline
        Branin & 20, 10 &  0.722 $\pm$ 0.1 & 0.543 $\pm$ 0.06 & - &
        \textbf{0.416 $\pm$ 0.01} \\
        \hline
        Hartmann6 & 20, 20 & -2.647 $\pm$ 0.13 & -2.672 $\pm$ 0.07 & - & \textbf{-3.158 $\pm$ 0.04} \\
        \hline
        Hartmann6 & 20, 10 & -2.231 $\pm$ 0.04 & -2.647 $\pm$ 0.13 & - & \textbf{-2.809 $\pm$ 0.04} \\
     \hline
     \multicolumn{6}{|c|}{Hyperparameter search at $100$ Epochs} \\
     \hline
      CIFAR-10 &  16, 100 & 92.66 & 92.82 &  92.93 & \textbf{93.62}\\
     \hline
      CIFAR-100 & 16, 100 & 69.23 & 69.23  &  69.82 & \textbf{71.66} \\
      \hline
     \multicolumn{6}{|c|}{Architecture search at $20$ Epochs} \\
     \hline
      CIFAR-10 & 80, 100 &  91.72 & 91.83  &  \textbf{92.72}  & 92.54 \\
     \hline
      CIFAR-100 & 80, 100 & 67.48 & 67.96 & \textbf{69.62} & 68.91 \\
     \hline
    \end{tabular}
\end{center}
\caption{ We compare the performance of SHAC with RS, RS-2X, and
  NAS{-PPO} on optimizing architectures,
  hyperparameters and synthetic functions on varying budgets. For
  synthetic functions, we report the mean across $5$ trials and the
  standard error.  For architecture and hyperparameter search, we
  report the mean of the top $5$ values at the end of the optimization
  procedure.} \label{tab:table1}}
\end{table*}

We cast hyperparameter search as black-box optimization, where the
objective function is the validation accuracy of a network trained
using a setting of hyperparameters. We fix the architecture to be
NASNET-A~\cite{nas2} with $9$ cells and a greatly reduced filter size
of $24$ for fast evaluation. We discretize the hyperparameters to be
able to utilize the NAS{-PPO} code directly. A
candidate point $x$ is a $20$D discrete vector that represents $20$
different hyperparameters including the learning rate, weight decay
rate for each cell, label smoothing and the cell dropout rate for each
cell. For the full specification of the search space, see the Appendix
\ref{hyper_search}. From the training sets of CIFAR-10 and CIFAR-100,
we set aside validation sets with $5000$ examples. Each black-box
evaluation $b(x)$ involves training NASNET-A for $100$ epochs on the
training set and reporting the validation accuracy. We allow $1600$
evaluations for RS, NAS{-PPO}, and SHAC, with batches of $100$
workers. We set the maximum number of classifiers in SHAC to $15$ and
the classifier budget to $100$ points per classifier.

\comment{
Since all the hyperparameters are discrete, we represent them
as ordinals in the $20$ dimensional feature representation provided
to the classifiers in SHAC.
}

We report the results in \tabref{tab:table1} and
\figref{fig:cifar_proxy}. SHAC significantly outperforms RS-2X and
NAS{-PPO} in these experiments. On CIFAR-10, SHAC obtains a $0.8\%$
gain over RS-2X and $0.69\%$ gain over NAS{-PPO} and on CIFAR-100, the
gain over RS and NAS{-PPO} increases to $2.42\%$ and $1.84\%$
respectively. One may be able to achieve better results using
NAS{-PPO} if they tune the hyperparameters of NAS{-PPO} itself,
\eg~learning rate, entropy coefficient, etc. However, in real black
box optimization, one does not have the luxury of tuning the
optimization algorithm on different problems. To this end, we did not
tune the hyperparameters of any of the algorithms and ran each
algorithm once using the default parameters. SHAC achieves a
significant improvement over random search without specific tuning for
different search spaces and evaluation budgets.

\subsection{Architecture Search} \label{arch_proxy} 

\begin{figure*}[t]
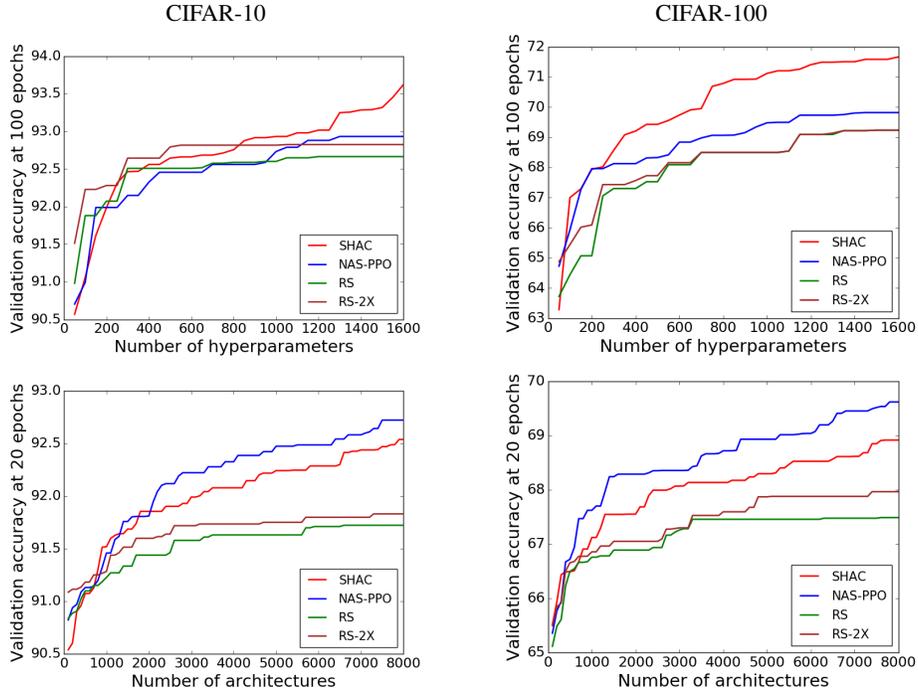

{
\small
\begin{center}
\begin{tabular}{@{}c@{\hspace*{1cm}}c@{\hspace*{.3cm}}c@{}}
CIFAR-10 & CIFAR-100 \\
\includegraphics[width=.4\textwidth]{cifar10_hparams_2X_proxy.pdf}
&
\includegraphics[width=.4\textwidth]{cifar100_hparams_2X_proxy.pdf}
& \\
\includegraphics[width=.4\textwidth]{cifar10_2X_proxy.pdf}
&
\includegraphics[width=.4\textwidth]{cifar100_2X_proxy.pdf}
& \\
\end{tabular}
\end{center}
}
\vspace*{-.1cm}
\caption{ We plot the mean of the top-$5$ scores obtained by SHAC,
  NAS-PPO and RS on CIFAR-10 and CIFAR-100 on both hyperparameter and
  architecture search via $1600$ and $8000$ evaluations respectively.}
\label{fig:cifar_proxy}
\vspace*{-.2cm}
\end{figure*}

We cast architecture search as black box optimization, where the
objective function is the validation accuracy of a given
architecture. For architecture search, a candidate point $x$ is a
$40$D discrete code that represents the design choices for a
convolutional cell. We follow the convolutional cell encoding proposed
in \cite{nas2} with minor modifications as communicated by the
original authors as outlined in Appendix~\ref{arch_search}. We use the
same validation split as in the hyperparameter search experiments
above. Each black box evaluation involves training an architecture
defined by a $40$D code for $20$ epochs on the training set and
reporting the validation accuracy. Each algorithm is provided with a
budget of $8000$ total evaluations computed in parallel using $100$
workers. For SHAC this means the $8000$ evaluations are evenly split
into $80$ synchronous rounds of $100$ evaluations. NAS is given an
advantage by allowing the algorithm to update the RNN parameters every
$5$ evaluations asynchronously.  This is a more generous budget
consistent with the conditions that NAS was designed for. For SHAC, we
set the maximum number of classifiers to $18$ and the minimum budget
per classifier to $400$.

\comment{We convert all of the features in the $40$D code that have no
  implicit ordering, such as the convolution type, into a one-hot
  representation. For the other features, we maintain an ordinal
  representation. In this way, the 40 dimensional feature vector
  became a 180 dimensional feature vector that could easily be handled
  by the gradient boosted trees. The configuration of NAS, the
  definition and training scheme of the convolutional networks
  themselves were kept at the latest defaults provided by the original
  authors and have some minor changes from their original
  publication. We document these minor changes in the Appendix.}

We report the results in \tabref{tab:table1} and
\figref{fig:cifar_proxy}. On CIFAR-10, SHAC demonstrates a gain of
$0.8\%$ and $0.7\%$ over RS and RS-2X while underperforming NAS-PPO by
$0.2\%$. On CIFAR-100, SHAC outperforms RS and RS-2X by $1.4\%$ and
$0.9\%$ respectively, but underperforms NAS-PPO by $0.7\%$. We note
that NAS-PPO is a complicated method with many hyperparameters that
are fine-tuned for architecture search on this search space, whereas
SHAC requires no tuning. Further, SHAC outperforms NAS-PPO on more
realistic evaluation budgets discussed for hyperparameter search
above. Finally, in what follows, we show that SHAC improves NAS-PPO in
terms of architecture diversity, which leads to improved final
accuracy when a shortlist of architectures is selected based on $20$
epochs and then trained for $300$ epochs.

\comment{As expected, for both SHAC and NAS-PPO, the mean of the top-5
  values gradually increase over the course of the optimization
  demonstrating their effectiveness as good black-box optimization
  strategies.}

\subsection{Two-stage Architecture Search}

\comment{
To apply architecture search in practice, we need
algorithms that find architectures that generalize well after they are
trained to convergence, not just for 20 epochs as in the section
above. 
}

To achieve the best results on CIFAR-10 and CIFAR-100, one needs to
train relatively wide and deep neural nets for a few hundred
epochs. Unfortunately, training deep architectures until convergence
is extremely expensive making architecture search with thousands of
point evaluations impractical. Previous work (\eg~\cite{geneticcnn}
and \cite{nas2}) suggests using a {\em two-stage} architecture search
procedure, where one adopts a {\em cheaper} proxy objective to select
a shortlist of top candidates. Then, the shortlist is evaluated on the
real objective, and the best architectures are selected. During proxy
evalution one trains a smaller shallower version of the architectures
for a small number of epochs to improve training speed. We follow the
proxy setup proposed by~\cite{nas2}, where the architectures are
trained for $20$ epochs first, as shown in~\figref{fig:cifar_proxy}
and \tabref{tab:table1}. Then, we select the top $50$ candidates based
on the proxy evaluation and train a larger deeper version of such
architectures for $300$ epochs, each $5$ times using different random
seeds.  We report the mean validation and test accuracy of the top $5$
architectures among the shortlist of $50$ for different algorithms in
\tabref{tab:table2}. Surprisingly, we find that all of the black box
optimization algorithms are competitive in this regime, with SHAC and
NAS achieving the best results on CIFAR-10 and CIFAR-100 respectively.

\begin{table*}[t]
{ \small
\begin{center}
  \begin{tabular}{| c | c | c | c | c | c | c |}
   \hline
       & \multicolumn{3}{c|}{Validation} & \multicolumn{3}{c|}{Test}\\
   \hline
   Dataset & RS & NAS{-PPO} & SHAC 
           & RS & NAS{-PPO} & SHAC \\
   \hline
   CIFAR-10 & 96.11 & 96.16 & {\bf 96.30}
            & 95.67 & 95.87 & {\bf 95.91} \\
   \hline
   CIFAR-100 & 79.06 & {\bf 79.53} & 79.37  
             & 79.59 & {\bf 79.93} & 79.80 \\
    \hline
  \end{tabular}
\end{center}
\vspace*{-.1cm}
\caption{ We select a shortlist of $50$ candidate architectures
  identified by each search algorithm based on their performance on
  the validation set after $20$ epochs. Then, we train each of these
  candidates for $300$ epochs and average their performance on the
  validation set across $5$ trials. We then report the mean performance of
  the top-$5$ architectures of each search algorithm on both the
  validation and test set, \ie~each number is an average across $25$ runs. \label{tab:table2}
 }
}
\label{tab:real}
\end{table*}

\begin{figure*}[t]
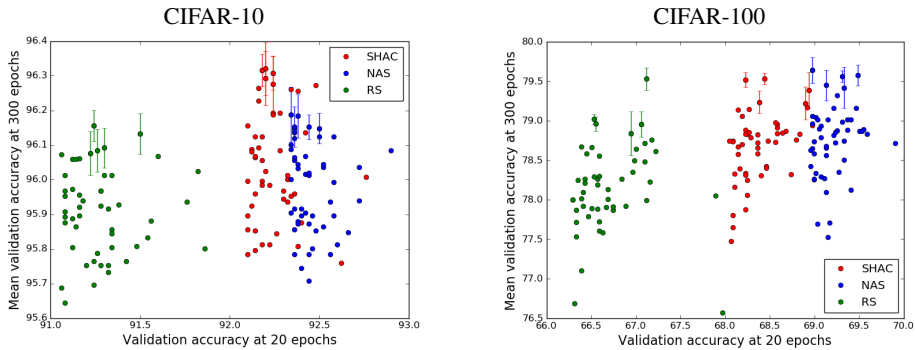

{
\small
\begin{center}
\vspace*{-.1cm}
\begin{tabular}{@{}c@{\hspace*{1cm}}c@{\hspace*{.3cm}}c@{}}
CIFAR-10 & CIFAR-100 \\
\includegraphics[width=.4\textwidth]{cifar10_scatter.pdf}
&
\includegraphics[width=.4\textwidth]{cifar100_scatter.pdf}
\end{tabular}
\end{center}
}
\caption{ We plot the validation accuracy at $300$ epochs ("real"
  objective) averaged across 5 runs \vs $20$ epochs ("proxy"
  objective) for the top $50$ architectures found by each search
  algorithm. Error bars are shown for the top $5$ architectures based
  on the metric. We observe that there is a weak correlation between
  the proxy and final metric.}
\label{fig:scatter}
\vspace*{-.2cm}
\end{figure*}

\comment{
\figref{fig:scatter}
plots the performance of these architectures on the real task as a
function of their performance on the proxy task, showing a weak
correlation between the proxy objective and real objective.
}

To investigate the correlation between the proxy and final objective
functions we plot the final measurements as a function of the proxy
evaluation in \figref{fig:scatter} for the shortlist of top $50$
architectures selected by each algorithm. We find that the correlation
between the proxy and the final metrics is not strong at least in this
range of the proxy values. When there is a weak correlation between
the proxy and the real objective, we advocate generating a {\em
  diverse} shortlist of candidates to avoid over-optimizing the proxy
objective. Such a {\em diversification} strategy has been extensively
studied in finance, where an investor constructs a diverse portfolio
of investments by paying attention to both expectation of returns and
variance of returns to mitigate unpredictable risks~\cite{portfolio}.

\begin{figure*}[t]
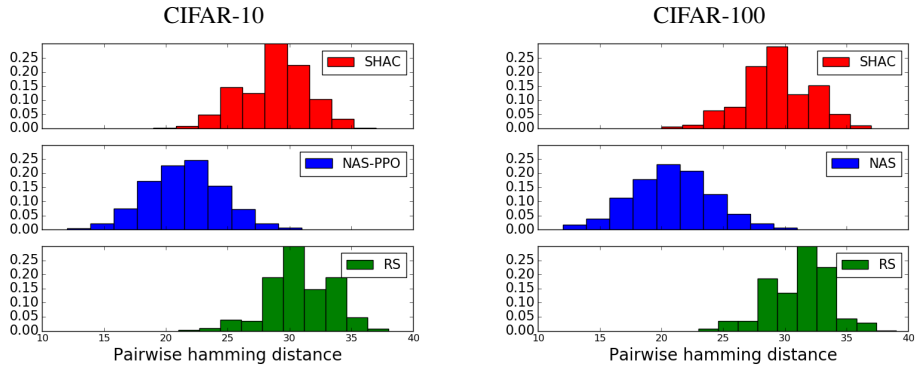

{
\small
\begin{center}
\begin{tabular}{@{}c@{\hspace*{1cm}}c@{\hspace*{.3cm}}c@{}}
CIFAR-10 & CIFAR-100 \\
\includegraphics[width=.4\textwidth]{cifar10_hist.pdf}
&
\includegraphics[width=.4\textwidth]{cifar100_hist.pdf}
\end{tabular}
\end{center}
}
\caption{We plot a histogram of the pairwise hamming distances of the
  shortlist of $50$ architectures found by SHAC, NAS-PPO, and RS. RS
  and SHAC promote diversity in the candidates. We find this behavior
  consistent across both CIFAR-10 and CIFAR-100.}
\label{fig:hamming}
\end{figure*}

Random search naturally generates a diverse candidate shortlist. The
cascade of $K$ classifiers in SHAC identifies a promising
${1}/{2^K}$th of the search space, which is still a large fraction of
the original high dimensional space. SHAC exhibits no preference
between the candidate points that make it past all of the classifiers,
hence it tends to generate a diverse candidate shortlist for large
search spaces. To study the shortlist diversity for different
algorithms, we depict the histogram of the pairwise Hamming distances
among the $40$D codes in the shortlist selected by each algorithm in
\figref{fig:hamming}. We approximate the distance between two $40$D
codes $x_i$ and $x_j$ representing two architectures via
$\sum_{n=1}^{40}\mathbbm{1}(x_i^n \neq x_j^n )$. This approximation
gives us a general sense of the diversity in the population, but maybe
improved using other metrics (\eg~\cite{kandasamy2018neural}). In
\figref{fig:hamming}, we observe that RS and SHAC have the most degree
of diversity in the candidate shortlists followed by NAS{-PPO}, the
mode of which is shifted to the left by $10$ units. Based on these
results we conclude that given the SHAC algorithm presents consistent
performance across differnt tasks and sample budgets
(\tabref{tab:table1} and \tabref{tab:table2}) despite its
deceptively simple nature.

\section{Conclusion}
We propose a new algorithm for parallel black box optimization called
SHAC that trains a cascade of binary classifiers to iteratively cull
the undesirable regions of the search space.  On hyperparameter search
with moderate number of point evaluations, we significantly outperform NAS-PPO and
RS-2X; random search with twice the number of evaluations.  On
architecture search, SHAC achieves competitive performance relative to
NAS-PPO, while outperforming RS-2X. SHAC is simple to implement and
requires no tuning of its own configuration parameters making it easy
to use for practitioners and a baseline for further research in
black-box optimization algorithms. Given the difficulty of
benchmarking architecture search algorithms, one should have a strong
bias towards algorithms like SHAC that are extremely simple to
implement and apply.

\section{Acknowledgements}

We would like to thank Barret Zoph and Quoc Le for providing their
implementation of the NAS-PPO and the convolutional cells used in
\cite{nas2}. Further, we would like to thank Azalia Mirhoseini, Barret
Zoph, Carlos Riquelme, Ekin Dogus, Kevin Murphy, and Jonathen Shlens
for their helpful suggestions and discussions at various phases of the
project.

\bibliography{paper}
\appendix

\section{Appendix}

\subsection{Hyperparameters for the convolutional networks} \label{arch_hyper}

We train each architecture for $20$ epochs on the proxy objective using Adam optimizer~\citep{adam}, a learning rate of $0.01$,
a weight decay of $10^{-4}$, a batch size of $256$, a filter size of $24$ and a cosine learning rate schedule.
The validation accuracy at $20$ epochs is used as the proxy metric.
We select a shortlist of $50$ architectures according to this proxy metric and train them with some \hyper changes to facilitate training longer and larger models.
The filter size of the shortlisted architectures are increased to $64$ and then are trained for $300$ epochs using SGD with momentum, a learning rate of $0.1$,
a smaller batch size of $64$, and a path dropout rate of $0.8$.
For the shortlisted architectures we plot the mean validation accuracy at $300$ epochs across $5$ runs.

\subsection{Search space for hyperparameter search} \label{hyper_search}

For hyperparameter search, we search over the learning rate, label smoothing, the dropout on the output activations of each of the 9 cells and the weight decay rate for each of the 9 cells thus obtaining a 20 dimensional search space. For each of these hyperparameters we search over the following possible values.

1. Label Smoothing - 0.0, 0.1, 0.2, 0.3, 0.4 and 0.5.

2. Learning rate - 0.0001, 0.00031623,  0.001, 0.01, 0.025, 0.04, 0.1,  0.31622777 and 1.

3. Weight decay rate - 1e-6, 1e-5, 5e-4, 1e-3, 1e-2 and 1e-1

4. Cell dropout - 0.0, 0.1, 0.2, 0.3, 0.4, 0.5, 0.6 and 0.7

\subsection{Search Space for architecture search} \label{arch_search}
For selecting architectures, we search over convolutional building blocks defined by \citet{nas2},
with the following minor modifications based on the communications with the respective authors: 
(1) We remove the $5\!\times\!5$ and $7\!\times\!7$ max pooling operations.
(2) We remove the option for choosing "the method for combining the hidden states" from the search space.
      By default, we combine the two hidden states by adding them.
With these modifications, a     cell is represented by a $40$ dimensional discrete code.

\end{document}